\documentclass[10pt]{article}

\usepackage{times}
\usepackage{epsfig}
\usepackage{graphicx}
\usepackage{amsmath}
\usepackage{amssymb}
\usepackage[preprint, nonatbib]{neurips_2019}
\usepackage{subcaption}
\usepackage{pgfplotstable}
\usepackage{booktabs} 
\usepackage{rotating}
\usepackage{multirow} 
\usepackage{array} 
\usepackage{makecell} 
\usepackage{pdfpages} 
\usepackage{adjustbox} 
\usepackage{tablefootnote} 
\usepackage{threeparttable} 
\usepackage[printonlyused,smaller,nohyperlinks]{acronym}
\graphicspath{ {images/} }

\usepackage[pagebackref=true,breaklinks=true,colorlinks,bookmarks=false]{hyperref}

\begin{document}
	
	\title{$\beta-$DVBF: Learning State-Space Models for Control from High Dimensional Observations}
	
	\author{Neha Das\\
			{\tt\small neha191091@gmail.com}\\
			\And 
			Maximilian Karl\thanks{Machine Learning Research Lab, Volkswagen Group, Munich, Germany}\\
			{\tt\small karlma@argmax.ai}
			\And 
			Philip Becker-Ehmck\footnotemark[1]\\
			{\tt\small philip.becker-ehmck@argmax.ai}
			\And 
			Patrick van der Smagt\footnotemark[1]
	}
	
	\maketitle

	\begin{center}
		\textbf{Disclosure}: Parts of this work have been submitted in form of a Master's Thesis towards partial fulfillment of the requirements for a Masters program at the Technical University of Munich \cite{NehaDas:Thesis:2019}	
	\end{center}

	\abstract
	Learning a model of dynamics from high-dimensional images can be a core ingredient for success in many applications across different domains, especially in sequential decision making. However, currently prevailing methods based on latent-variable models are limited to working with low resolution images only. In this work, we show that some of the issues with using high-dimensional observations arise from the discrepancy between the dimensionality of the latent and observable space, and propose solutions to overcome them.
	\endabstract

	\section{Introduction}
	Learning a probabilistic model for sequential data is a key step towards solving a lot of interesting problems, including analysis and deconstruction of auditory sequences \cite{turner2010statistical}, predicting the next piece of information given previously recorded data such as video frames \cite{jayaraman2018time} and text \cite{xie2017neural}, and controlling an agent to perform specific tasks (model-based reinforcement learning \cite{deisenroth2011pilco}). While in this paper, we consider probabilistic models especially tuned for the needs of the last, i.e with control inputs, the approaches we discuss can be applied to control-less environments as well. 
	
	A key feature required in control-based models is that they should be able to generate a feasible trajectory distribution given a control policy.	To this end, one of the more successful solutions proposed in the past for modeling dynamical systems is \ac{DVBF} \cite{karl2016deep} - a framework for learning a State-Space Model given sequential observations from the environment in an unsupervised manner. Although models learned via \ac{DVBF} have shown exceptional predictive performance on low dimensional data, they lose this capability when scaled to high-dimensional observations.  In this work, we conclude that the reason for this discrepancy lies in the relative difference between the dimensions of observation and latent-space. We propose an extension to the \ac{DVBF} model that takes this relative difference into account, thus improving its scalability. In the experimental section, we further discuss how the different constructs introduced in \ac{DVBF} and in this work assist in learning a model with good prediction capabilities and to which degree. We validate these discussions by showcasing results on a number of different data sets.
	
	\section{Background and Related Works}
	
	We consider a dynamical system that is controlled via a sequence of known control inputs $u_{1:T-1}$ and is only observable through a sequence of images $x_{1:T}$ obtained for a time horizon of length $T$. Since images, while being high dimensional fail to capture the exact system state at each point of time in a Markovian fashion, it is natural to model the generative distribution using the State Space Models framework. Formally, the probabilistic model governing the observation distribution through the \ac{SSM} can be described as:
	
	\begin{equation}\label{eq:SSM}
	P_\theta(x_{1:T}|u_{1:T}) = \int P_{\theta_0}(x_1, z_1)\prod_{i=2}^{T}P_{\theta_{t}}(z_t|z_{t-1}, u_{t-1})P_{\theta_{e}}(x_t|z_t) dz_{1:T}
	\end{equation}
	
	where $z_{1:T}$ denote the latent states for each time step $t \in \{1, 2, ... , T\}$ and $\theta = \{\theta_0, \theta_t, \theta_e\}$ represent the parameters of the initial, transition and emission distributions.
	
	Our task then, becomes two-fold: \textbf{(a)} Identify the system dynamics, i.e the transition model $p(z_t|z_{t-1}, u_{t-1})$ and the emission model $p(x_t|z_t)$ from the data and control sequence alone, and \textbf{(b)} Identify the current system state given the sequence of previous observations and controls, i.e $p(z_t|x_{1:t}, u_{1:t-1})$. This is often known as the filtering distribution, is useful for applications that need to predict the future given the past data.
	
	A number of previous approaches \cite{bayer2014learning, 2015deepkalmanfilters, chung2015recurrent, karl2016deep, fraccaro2017disentangled, denton2018stochastic, lee2018stochastic} have used a framework based on sequential \ac{VAE} in an attempt to work towards both these goals. The goal here is to maximize the evidence lower bound ($\mathcal{L}_{ELBO}$) to the data log likelihood in order to learn a manifold $Z$ in a lower dimensional space to which the data $X$ can be projected for effective compression. More specifically, we want to learn a  probabilistic approximation of the said projection - i.e posterior distribution $p(z|x)$, as well as the generative distribution $p(x|z)$ for the data given its corresponding latent projection. 
	
	Following this paradigm, the $\mathcal{L}_{ELBO}$ for sequential data $x_{1:T}$ can be written as:
	\begin{equation}\label{eq:VI}
	\begin{split}
	\mathcal{L}_{ELBO} &= \int q_\phi(z_{1:T}|x_{1:T},u_{1:T-1}) \log \frac{p_\theta(x_{1:T}|z_{1:T}, u_{1:T-1}) p_\theta(z_{1:T}|u_{1:T-1})}{q_\phi(z_{1:T}|x_{1:T}, u_{1:T-1})} dz_{1:T} \\
	&= \underbrace{\underset{z_{1:T} \sim {q_\phi}}{\mathbb{E}} \big[\log p_\theta(x_{1:T}|z_{1:T}, u_{1:T-1})\big]}_{\color{red} \underset{\scriptstyle error}{reconstruction}}- KL(\underbrace{q_\phi(z_{1:T}|x_{1:T}, u_{1:T-1})}_{\color{red} \underset{\scriptstyle model}{\small recognition}}||\underbrace{p_\theta(z_{1:T}|u_{1:T-1})}_{\color{red} \underset{\scriptstyle}{prior}}) \\
	&\leq \log p(x_{1:T}|u_{1:T-1}) 
	\end{split}
	\end{equation}
	
	where $q_\phi$ denotes the approximate posterior distribution over the latent variables $z_{1:T}$ given the data $x_{1:T}$ and control $u_{1:T-1}$ inputs, $\phi$ denotes the parameters for said distribution and $\theta$ denotes the parameters for the generative distribution. 
	The maximization of $\mathcal{L}_{ELBO}$ with respect to $\phi$ and $\theta$ can therefore facilitate learning a function to infer meaningful latents from the data sequence and a mapping from the latent space to the observation space.
	
	To ensure that the state at each time-step $t$ ($z_t$) reflects the complete system information, several of these methods \cite{2015deepkalmanfilters, fraccaro2017disentangled, karl2016deep} further encode \ac{SSM} assumptions (equation \ref{eq:SSM}) in the generative model leading to the following formulation for $\mathcal{L}_{ELBO}$
	
	\begin{equation} \label{eq:seq_ELBO}
	\begin{split}
	\mathcal{L}_{ELBO}
	&= \underset{z_{1:T} \sim {q_\phi}}{\mathbb{E}} \big[\sum_{t=1}^T \log \underbrace{ p_{\theta_e}(x_{t}|z_{t})}_{\color{red}\underset{\scriptstyle model}{emission}}\big] - KL(\underbrace{q_\phi(z_{1:T}|x_{1:T}, u_{1:T-1})}_{\color{red} \underset{\scriptstyle model}{\small recognition}}||p_{\theta_0}(z_{1})\prod_{t=1}^{T-1}\underbrace{p_{\theta_t}(z_{t+1}|z_{t}, u_{t})}_{\color{red}\underset{\scriptstyle model}{transition}})
	\end{split}
	\end{equation}	
	
	Instead of designing the reconstruction error and the prior distribution, our task now focuses on  the specification of emission and transition models in addition to the recognition model. These are often modeled as conditional Gaussian distributions for simplicity. In case of the transition distribution, many works also opt for a more structured form such as Locally Linear Dynamics \cite{watter2015embed, karl2016deep} and \ac{SLDS} \cite{becker2018switching} in a bid to reduce the size of the parameter space and include some prior knowledge regarding the system dynamics. 
	
	The recognition model can be similarly simplified by first decomposing it auto-regressively and subsequently imposing \ac{SSM} assumptions from equation \ref{eq:SSM} as before:
	\begin{equation}
	q(z_{1:T}|x_{1:T}, u_{1:T-1}) = q_{\phi_0}(z_1|x_{1:T}) \prod_{i=1}^{T-1} q_\phi(z_{t+1}|z_{t}, u_{t}, x_{t:T})
	\label{eq:posterior_decomp_2}
	\end{equation}
	
	The probabilistic factor $q_\phi(z_{t+1}|z_t, u_t, x_{t:T})$ is an approximation of the smoothing distribution $p(z_t|x_{1:T}, u_{1:T-1})$. In case of online inference, where future observations are unavailable to us, one can also replace it with an approximation to the filtering distribution $p(z_t|x_{1:t}, u_{1:t-1})$ - i.e $q_\phi(z_{t+1}|z_t, u_t, x_{t})$. As with the emission and transition distributions, for simplicity, these distributions can be modeled via Gaussian PDFs, with the parameters of the densities being the output of a recurrent or bi-directional neural network \cite{2015deepkalmanfilters}.
	
	\subsection{Deep Variational Bayes Filter} \label{subsec:DVBF}
	
	For many cases however, the above formulation is not enough to facilitate learning a generative model with good predictive capabilities. Note that the generative model figures in the $\mathcal{L}_{ELBO}$ as the prior transition distribution. However, as pointed out by \cite{karl2016deep}, since this term is only a part of the KL divergence term in equation \ref*{eq:seq_ELBO}, it cannot directly use the gradients obtained from the emission error to update itself. Consequently, while the model learns to efficiently compress a given sequence, prediction is still a hard task. \cite{karl2016deep} demonstrates this by comparing results for modeling pendulum data for different architectures. In particular, Deep Kalman Filter \cite{2015deepkalmanfilters}, a model that uses the formulation and posterior decomposition discussed above is shown to learn only a part of the true state space. 
	
	Deep Variational Bayes Filter \cite{karl2016deep, karl2017unsupervised}, an architecture that forms an the essential core to our approach on the other hand, uses a two pronged approach to deal with this predilection:
	
	\begin{itemize}
		\item The approximate filtering distribution $q(z_t|z_{t-1}, u_{t-1}, x_t)$ is restructured as the product of an inverse measurement and a posterior transition distribution:
		\begin{equation}\label{eq:posterior_decomp_3}
		q_\phi(z_t|z_{t-1}, u_{t-1}, x_t) \propto \underbrace{q_{\phi_e}(z_t|x_t)}_{\color{red} \underset{{\scriptstyle measurement}}{inverse-}} \times \underbrace{q_{\phi_t}(z_t|z_{t-1}, u_{t-1})}_{\color{red} \underset{{\scriptstyle transition}}{posterior}}
		\end{equation}
		This is inspired from (but is not equivalent to) the Bayesian belief update \cite{murphy2012machine}, which is a theoretically grounded mechanism for updating prior beliefs with current measurements. Breaking down the approximate filtering distribution (and in essence the recognition model) this way enables its explicit regularization with the transition prior. As we show in the experimental section, even in the absence of the second step, this seems to induce essential support for learning system dynamics.
		\item Distribution parameters are shared partially or fully between the prior and posterior transition models - $p_{\theta_t}(z_t|z_{t-1}, u_{t-1})$ and $q_{\phi_t}(z_t|z_{t-1}, u_{t-1})$. In \cite{karl2017unsupervised}, this translates to sharing the mean between the two distributions (modeled as Gaussian PDFs). This allows gradients from the reconstruction loss to flow back through the transition in addition to the gradients from the KL divergence term, thus inducing a stronger bias towards learning a good generative model.
	\end{itemize}
	
	\section{Method}\label{sec:Method}

	While in practice, \ac{DVBF} and similar approaches (e.g \cite{fraccaro2017disentangled}) are able to learn a dynamics model from low-dimensional image sequences, they may not perform well when the observation size is scaled to a higher resolution. We show this to be the case for atleast one of the above models (\ac{DVBF}) in the Experimental section.
	
	In order to introduce a method to alleviate this issue however, it is beneficial to first examine what causes it. A majority of the methods we have referred to here can all be summarized as the optimization of variational lower bound ($\mathcal{L}_{ELBO}$) to the data-log likelihood. Recall from equation \ref{eq:VI}, that $\mathcal{L}_{ELBO}$ contains the sum of two terms - the reconstruction error for the observed data, and the KL Divergence between the prior and the approximate posterior distributions over the latent space. When processing high-dimensional images, these two terms may have significantly different magnitudes owing to the fact that one of them - the KL divergence is calculated over the latent space and has been noted to have a relatively smaller contribution (magnitude-wise) than the other - the reconstruction loss, which is calculated over the higher-dimensional observation space. 
	
	Since the generative model is solely a part of the latter term, it is important that equal emphasis is provided to both terms while training. Therefore, while the models discussed above show good results for low dimensional images (where this quality  isn't lacking), may not perform well when the observation size is scaled to a higher resolution (where KL receives far less attention compared to the reconstruction loss by the training procedure).
	
	One measure, therefore, in order to provide greater emphasis on learning a good prior transition in addition to properly regularizing the recognition model, is to scale the KL divergence term with a new hyper-parameter $\beta \geq 1$, thus deriving a lower bound to $\mathcal{L}_{ELBO}$:
	
	\begin{equation}
	\begin{split}
	\mathcal{L}_{ELBO} \geq& \underset{z_{1:T} \sim {q_\phi}}{\mathbb{E}} \big[\log\underbrace{ p_\theta(x_{1:T}|z_{1:T}, u_{1:T})}_{\color{red}emission}\big] \\
	& -\beta \cdot KL\big[\underbrace{q_\phi(z_{1:T}|x_{1:T}, u_{1:T})}_{\color{red}recognition}||\underbrace{p_\theta(z_{1:T}|u_{1:T})}_{\color{red}prior\text{ }transition}\big]
	\end{split}
	\label{eq:LELBO_beta}
	\end{equation}
	
	In their work introducing $\beta$-VAEs, Higgins et al. use a similar formulation to learn disentangled factors underlying the observed (non-sequential) data. To do so, they assume a standard Normal prior over the latents and re-frame the problem of learning a generative model for the observations as constrained optimization - more specifically, the maximization of $\underset{z \sim q(z|x)}{\mathbb{E}}\big[ \log p(x|z)\big]$ subject to $KL(q(z|x)||p(z) \leq \epsilon)$. This is equivalent to optimizing the Lagrangian (under KKT conditions):
	
	\begin{equation}\label{eq:beta_ELBO}
	\begin{split}
	&\underset{z \sim q(z|x)}{\mathbb{E}}\big[ \log p(x|z)\big] - \beta \big(KL(q(z|x)||p(z)) - \epsilon\big)\\
	& \geq \underset{z \sim q(z|x)}{\mathbb{E}}\big[ \log p(x|z)\big] - \beta \big(KL(q(z|x)||p(z))\big) =\beta\mathcal{L}_{ELBO} \leq \mathcal{L}_{ELBO}
	\end{split}
	\end{equation}
	
	As a consequence of the constraint on the KL term, the approximate posterior can only diverge from the prior maximally by the amount $\epsilon$ which, if small, in practice encourages it to be factorized or diagonal, thus forcing the latent dimensions to be disentangled or uncorrelated. In \cite{higgins2017beta}, $\beta$ is treated as a hyper-parameter, implicitly determining the value of the constraint $\epsilon$. While in our case, we do not particularly care for disentangling the latent space, we employ the same concept of constrained optimization to force the approximate posterior and the prior terms closer. Choosing an appropriately high value of $\beta$ translates to picking a small value for $\epsilon$ implicitly, which in turn facilitates the learning of a good generative model which is also incorporated in the inference process. This is demonstrated in the experimental section for several different environments. 
	
	Our experiments additionally show that the optimum amount of $\beta$ is roughly around $\frac{dim_{x}}{dim_{z}}$, where $dim_{x}$ and $dim_z$ respectively signify the observation and latent dimensions (See Appendix). We noted while conducting experiments however, that a higher value of $\beta$ must be complemented with an appropriate annealing scheme to ensure a smooth training process. We therefore anneal the KL-divergence in our formulation of the $\mathcal{L}_{ELBO}$ while training by multiplying it with a factor that linearly increases from $0$ to the chosen $\beta$ value with the training time.
	
	It should be mentioned that Lee et al. also present a very similar formulation to ours in their work Stochastic Adversarial Video Prediction \cite{lee2018stochastic}. They do not, however, impose \ac{SSM} assumptions on the generative model or discuss why the coefficients $\lambda_1$ and $\lambda_{KL}$ multiplied with the reconstruction and KL terms respectively are needed, along with a more principled way to tune these parameters, as we showed in this work. 
	
	\section{Experiments}
	
	In this section, we apply the methods proposed above on two different non-Markovian systems and compare the quality of the resultant generative model with those resulting from previous approaches. We additionally analyze and evaluate the conditions for learning a good generative model, specifically in terms of the approximate posterior's chosen form and structure. Our analysis shows that explicitly decomposing the approximate posterior density in the manner shown in equation \ref{eq:posterior_decomp_3} turns out to be a key factor in learning a good model for transition and shows superior results for prediction tasks compared to approaches that infer from the observations and previous latents jointly using a single \ac{MLP} (or any other structure). 
	
	We additionally show that along with an appropriate $\beta$ value, this model produces results comparable to systems that additionally use more sophisticated constructs for modeling transitions (see section ) and/or share parameters between the prior and posterior transition distributions.

	\subsection{Pendulum}
	
	The Pendulum Dataset comprises of image sequences ($64 \times 64$ pixels) of a moving pendulum, represented using a Gaussian blob, along with corresponding  control inputs. The positions of the pendulum mass are sampled from a simulated environment introduced in \cite{karl2016deep} (See Appendix \ref{app:pend_dynamics} for details).
	
	For our analysis, we train four distinct models (see table \ref{tab:models}) - DVBF-SLDS \cite{karl2017unsupervised, becker2018switching} and Deep-Kalman-Filter \cite{2015deepkalmanfilters} serve as the two baselines that we compare our models against. DVBF-Non-Shared serves to show that the model performance is comparable to that of DVBF-SLDS even without a more sophisticated transition model and parameter sharing, as long as the approximate posterior is decomposed according to equation \ref{eq:posterior_decomp_3}.
	
	\begin{table}[h!]
		\centering
		\begin{center}
			\begin{tabular}{l c p{3.5cm} l p{1.5cm}}
				\toprule
				\bf Model & \bf Transition Network & \bf Shared Prior and Posterior parameters & \bf $\beta$ & \bf Annealing Temp\\ 
				\midrule
				DVBF-SLDS & \ac{SLDS} & mean $\mu_{trans}$ & $1.0$ & $1.0$\\
				DVBF-Non-Shared & \ac{MLP} & None & $1.0$ & $1.0$\\
				DVBF-Non-Shared-$\beta$ & \ac{MLP} & None & $8.0$ & $5e3$\\
				Deep-Kalma-Filter & Locally Linear & None & $1.0$ & 1.0\\
				\bottomrule
			\end{tabular}
		\end{center}
		\caption{\label{tab:models} Models trained on the Pendulum Environment}
	\end{table}	

\begin{figure}[h!]
		\centering
		\begin{tabular}{ccccc}
			\toprule
			&\bf DVBF-SLDS & \bf DVBF-Non-Shared & \bf DVBF-Non-Shared-$\beta$ & \bf Deep-Kalman-Filter\\ 
			\midrule
			\raisebox{-.5\normalbaselineskip}[0pt][0pt]{\rotatebox[origin=c]{90}{\hspace{2cm}\textbf{Angle}}} &
			{\includegraphics[width = 0.7in]{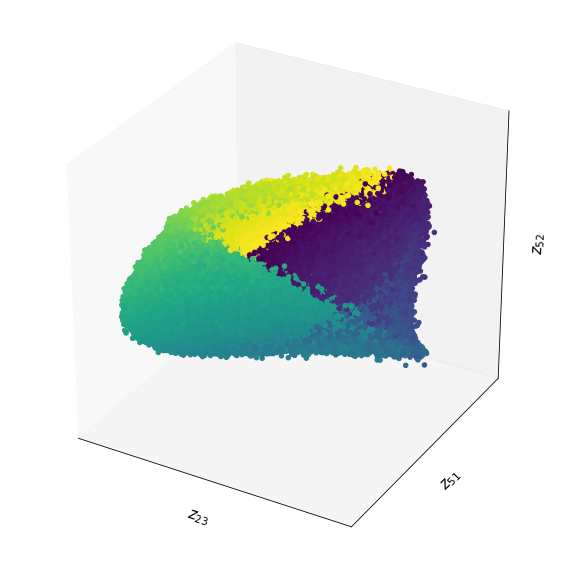}} &
			{\includegraphics[width = 0.7in]{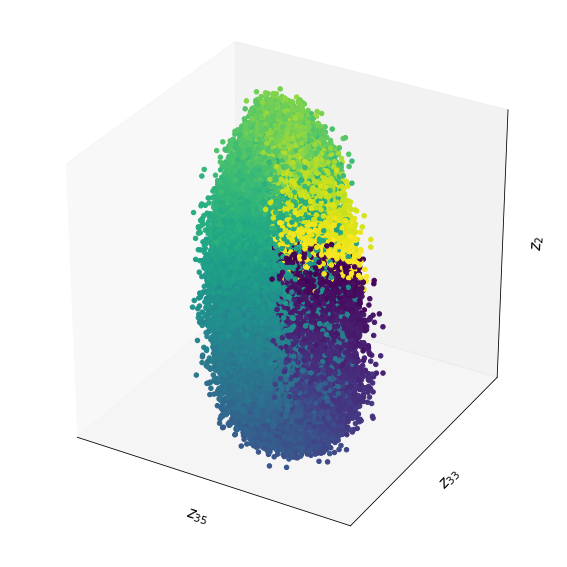}} &
			{\includegraphics[width = 0.7in]{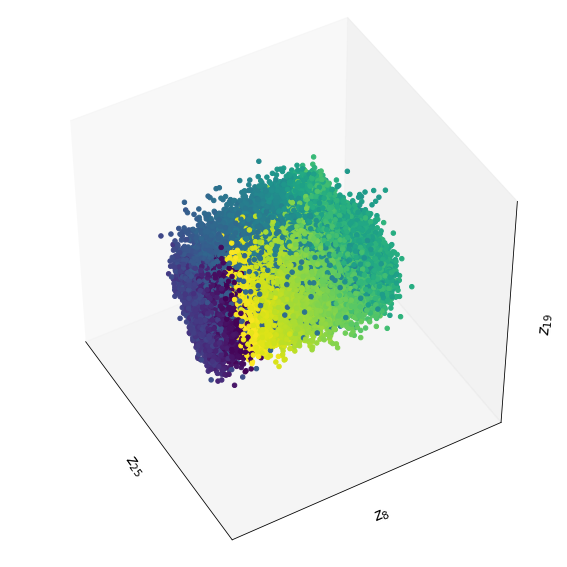}} &
			{\includegraphics[width = 0.7in]{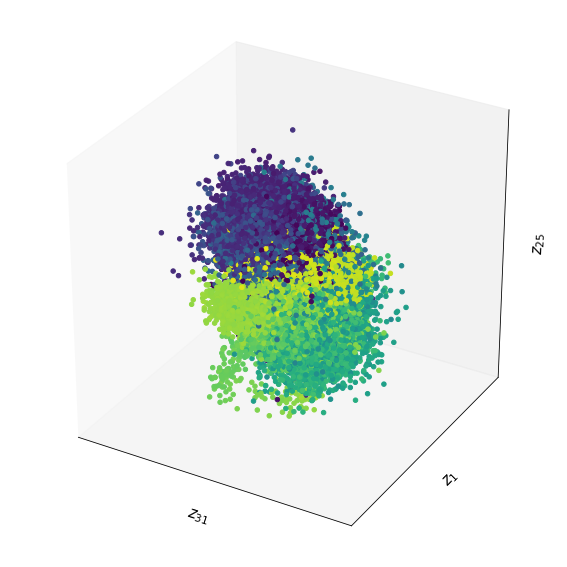}}\\
			\begin{turn}{90}
				\bf \hspace{0.2cm}Velocity
			\end{turn} &
			{\includegraphics[width = 0.7in]{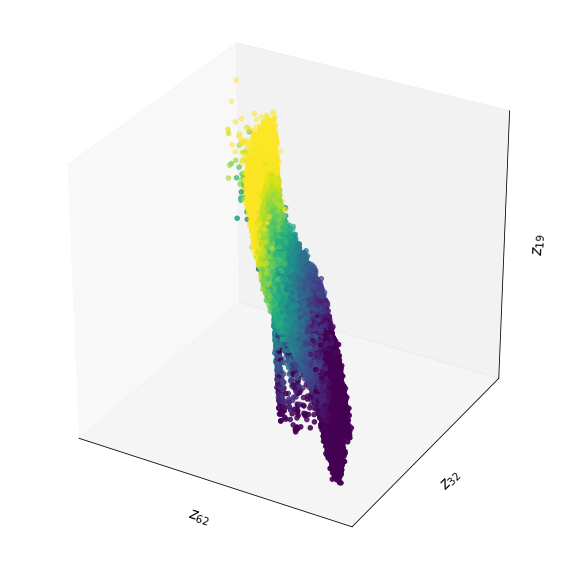}} &
			{\includegraphics[width = 0.7in]{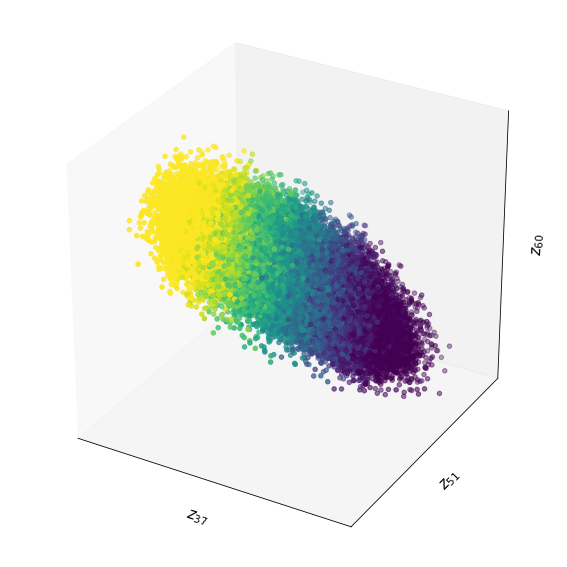}} &
			{\includegraphics[width = 0.7in]{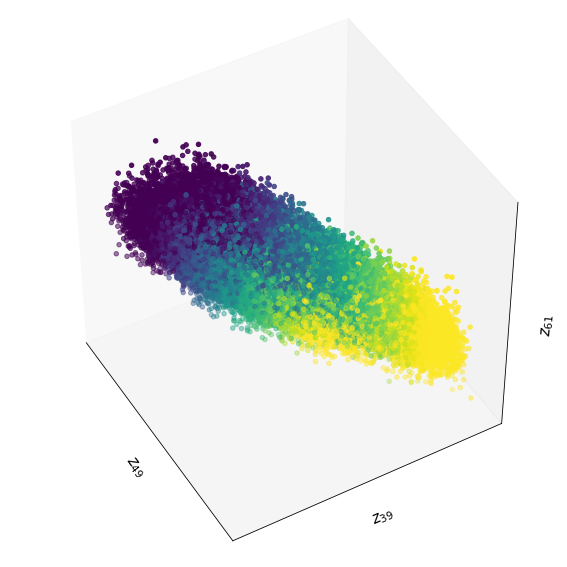}} &
			{\includegraphics[width = 0.7in]{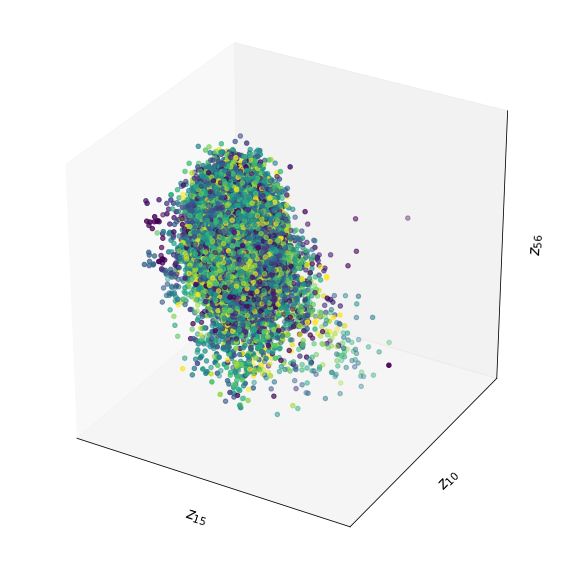}}
		\end{tabular}
		\caption{We color the three most correlated latent dimensions learned in the Pendulum Environment by ground truth values - pendulum angle $\psi$ and velocity $\dot{\psi}$. While Deep-Kalman-Filter fails to learn a latent dimension that encodes angular velocity, the rest of the models succeed on account of their decomposed approximate posteriors.}
		\label{fig:pend_latents}
	\end{figure}

	We use several measures to evaluate the quality of our model.
	Since the system state in this environment can be fully specified by the angle $\psi(t)$ and angular velocity $\dot{\psi}(t)$ of the pendulum mass at any given time, one of these measures is to check the whether the learned latents $z$ can correlate to these quantities (See Figure \ref{fig:pend_latents}). We further compare the models using other standard metrics established by previous works, such as the mean-squared error (MSE) \cite{becker2018switching} at varying time-steps as well as $\mathcal{L}_{ELBO}$, \ac{KL} divergence and \ac{NLL} in table \ref{tab:metrics}.
	
	\begin{table}[h!]
		\centering
		\begin{center}
			\begin{tabular}{l l l l p{1cm} p{1cm} p{0.7cm} p{0.7cm} p{0.7cm}}
				\toprule
				\bf Model & \bf $\mathcal{L}_{ELBO}$ & \bf KL & \bf NLL & \multicolumn{2}{c}{\bf Correlation} & \multicolumn{3}{c}{\bf MSE (For Steps)}\\ 
				& & & & \bf Angle & \bf Velocity  & \bf 1 & \bf 5  & \bf 10 \\ 
				\midrule
				DVBF-SLDS & $743.2$ & ${0.879}$ & $\mathbf{-749.8}$ & $0.70$ & $\mathbf{0.97}$ & $\mathbf{0.244}$ & $\mathbf{0.461}$ & $\mathbf{0.630}$ \\
				DVBF-Non-Shared & $\mathbf{746.4}$ & $1.828$ & $-748.2$ & $\mathbf{0.74}$ & $0.73$ & $0.245$ & $0.467$ & $0.641$ \\
				DVBF-Non-Shared-$\beta$ & ${746.1}$ & $\mathbf{0.774}$ & $-746.9$ & ${0.71}$ & $0.88$ & ${0.246}$ & $0.466$ & $0.631$ \\
				Deep-Kalman-Filter & $732.9$ & $9.825$ & $-742.8$ & $0.54$ & $0.10$ & $0.251$ & $0.731$ & $0.930$ \\
				\bottomrule
			\end{tabular}
		\end{center}
		\caption{\label{tab:metrics} The different metrics evaluated on models from  (\ref{tab:models}) trained on the Pendulum Environment.}
	\end{table}
	
\begin{figure}[h!]
		\centering
		\begin{tabular}{c}
			
			\midrule
			\multicolumn{1}{c}{\textbf{DVBF-SLDS}} \\
			\midrule
			{\includegraphics[width = 4in]{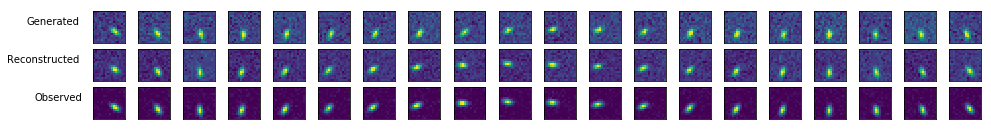}}\\
			\midrule
			\multicolumn{1}{c}{\textbf{DVBF-Non-Shared}} \\
			\midrule
			{\includegraphics[width = 4in]{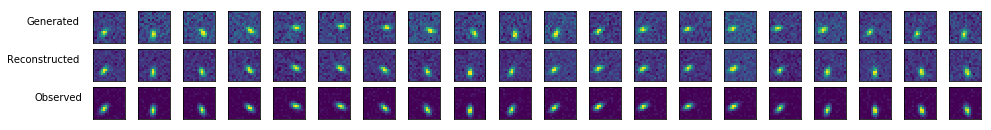}}\\
			\midrule
			\multicolumn{1}{c}{\textbf{DVBF-Non-Shared-$\beta$}} \\
			\midrule
			{\includegraphics[width = 4in]{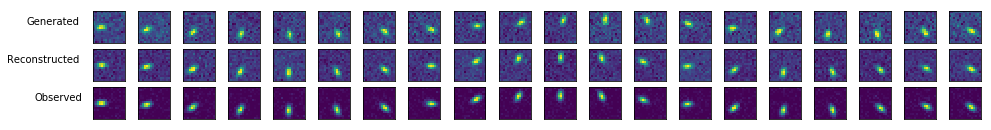}}\\
			\midrule
			\multicolumn{1}{c}{\textbf{Deep-Kalman-Filter}} \\
			\midrule
			{\includegraphics[width = 4in]{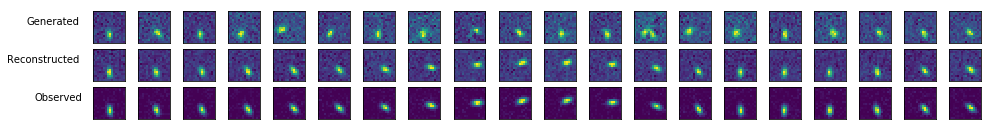}}\\
			\bottomrule
		\end{tabular}
		\caption{We show the generated, reconstructed and original trajectories obtained in the Pendulum Environment for the various models described in table \ref{tab:models}}.
		\label{fig:pend_results}
	\end{figure}

	Finally, we also qualitatively evaluate our models by visually comparing the generated and reconstructed trajectories with the original trajectory in Figure \ref{fig:pend_results}. For each such result in different environments (i.e \ref{fig:pend_results}, \ref{fig:vizdoom_results} and \ref{fig:bb_results} (In Appendix)), we show the generated trajectory where the latent vector at the first time step is sampled from the initial network and the next 39 latents are generated from the prior transition model and then decoded into an image. We show every second image from this sequence in the interest of brevity. We similarly show the reconstructed sequences, where the latent vector for each time-step is sampled from the approximate posterior distribution and then fed to the decoder to generate the reconstructed image sequence.
	
	As has been shown previously in \cite{karl2016deep}, Figure \ref{fig:pend_latents} demonstrates that Deep-Kalman-Filter \cite{2015deepkalmanfilters} is unable to model the angular velocity $\ddot{\psi}$. \ac{DVBF} attributes this to insufficient regularization of the approximate posterior with the prior and corrects this with posterior decomposition, parameter sharing and using structured transitions (See the results for DVBF SLDS). However, note that DVBF-Non-Shared also manages to learn both angular velocity and the pendulum angle with sufficient accuracy. This is indicative of the relative significance of the decomposition step for the filtering distribution with respect to additional measures such as incorporating structured transition dynamics and/or parameter sharing between the prior and posterior distributions. We conjecture that this step restricts the solution-space for possible inference models by inducing prior knowledge regarding the explicit encoding of the transition dynamics in the recognition model analogous to that in Recursive Bayesian Filter.
	
	Further, note from table \ref{tab:metrics} that a high $\beta$ value (DVBF-Non-Shared-$\beta$) may help provide adequate regularization when the model lacks other DVBF characteristics such as parameter-sharing and structured transition models.
	
	\subsection{Vizdoom}
	
\begin{figure}[h!]
		\centering
		{\includegraphics[width = 4in]{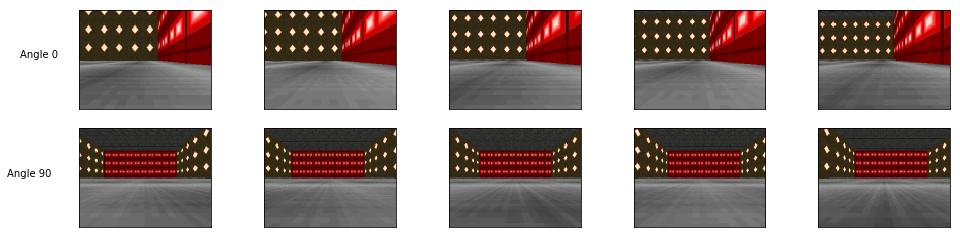}}
		
		\caption{An example sequence from the recorded observations in the Vizdoom Experiment.}
		\label{fig:vizdoom_data}
	\end{figure}

	This experiment focuses on the significance of a higher $\beta$ value when the corresponding images are high-dimensional. The dynamics of this environment are simulated using the pybox2d engine and imitate an agent (ball) rolling around in a bounded box. However, the images ($96\times128\times3$) in the data-sequence are agent-perspective views of the environment from two different angles in addition to the corresponding control vector (The forces in x and y directios) rendered using Vizdoom. To simplify the problem by a notch, we ensure that the agent's position is at any given time is observable by coloring the walls with repeated regular patterns shown in Figure \ref{fig:vizdoom_data}.
	
	We compare evaluation metrics for three models (See Table \ref{tab:vizdoom_models}) in this setting - DVBF-SLDS (DVBF with Switching Dynamics transition) is used as a baseline, DVBF-SLDS-$\beta$, which illustrates the clear advantage of having an appropriately high $\beta$ value for experiments that have a large difference between the size of the observation and latent space, and DVBF-Non-Shared-$\beta$, which is similar to its counterpart described in the previous experiments.
	
	\begin{table}[h!]
		\centering
		\begin{center}
			\begin{tabular}{l p{2cm} p{3.5cm} p{0.5cm} p{2cm}}
				\toprule
				\bf Model & \bf Transition Network & \bf Shared Prior and Posterior parameters & \bf $\beta$ & \bf Annealing Temperature \\ 
				\midrule
				DVBF-SLDS & \ac{SLDS} & mean $\mu_{trans}$ & $1.0$ & $1.0$ \\
				DVBF-SLDS-$\beta$ & \ac{SLDS} & mean $\mu_{trans}$ & $128$ & $1.5e3$\\
				DVBF-Non-Shared-$\beta$ & \ac{MLP} & None & $256$ & $1.5e3$\\
				\bottomrule
			\end{tabular}
		\end{center}
		\caption{\label{tab:vizdoom_models} Models trained on the Vizdoom Environment.}
	\end{table}
	
\begin{figure}[h!]
		\centering
		\begin{subfigure}[h]{8cm}
			\begin{tabular}{c p{1.7cm} p{1.7cm} p{1.7cm}}
				\toprule
				&\bf DVBF-SLDS & \bf DVBF-SLDS-$\beta$ & \bf DVBF-Non-Shared-$\beta$\\ 
				\midrule
				\raisebox{-.5\normalbaselineskip}[0pt][0pt]{\rotatebox[origin=c]{90}{\hspace{2cm}\textbf{Latent}}} &
				{\includegraphics[width = 0.7in]{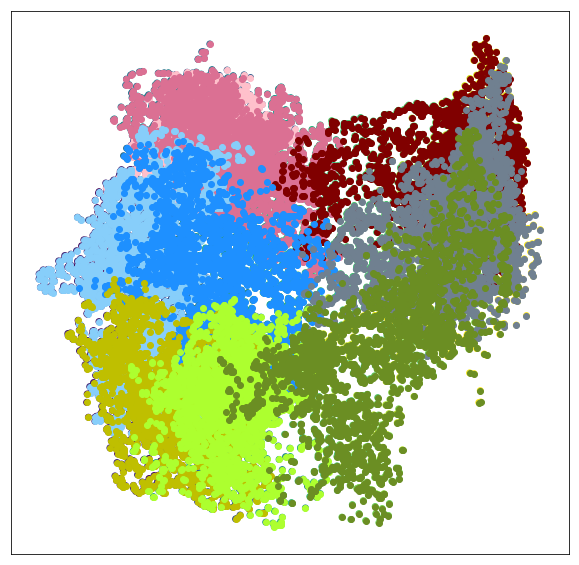}} &
				{\includegraphics[width = 0.7in]{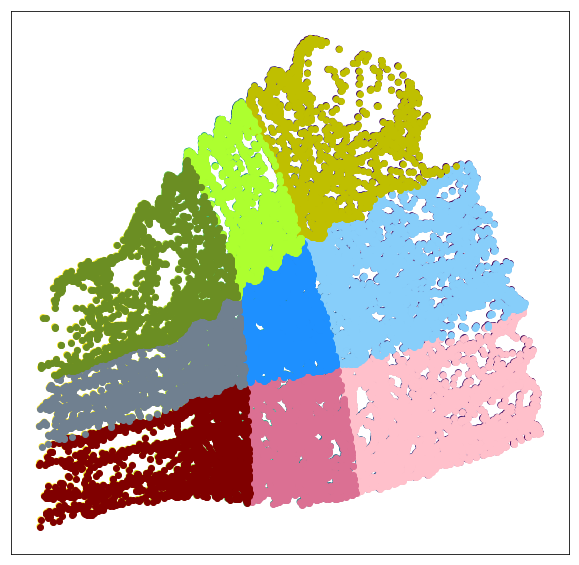}} &
				{\includegraphics[width = 0.7in]{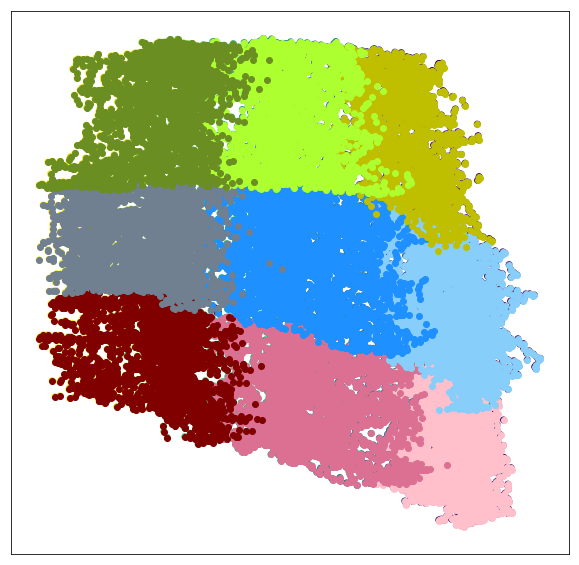}}\\
				\begin{turn}{90}
					\bf Velocity(x)
				\end{turn} &
				{\includegraphics[width = 0.7in]{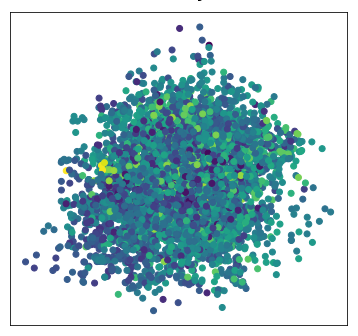}} &
				{\includegraphics[width = 0.7in]{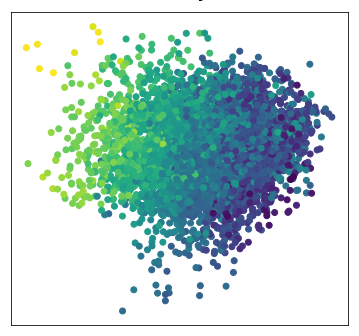}} &
				{\includegraphics[width = 0.7in]{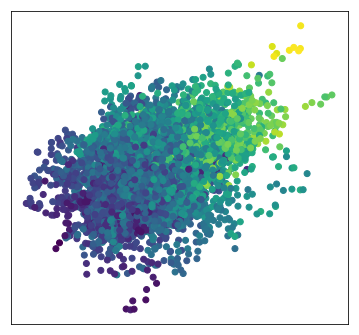}}\\
				\begin{turn}{90}
					\bf Velocity(y)
				\end{turn} &
				{\includegraphics[width = 0.7in]{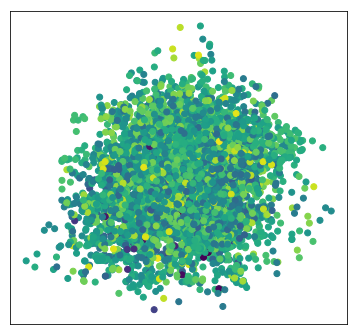}} &
				{\includegraphics[width = 0.7in]{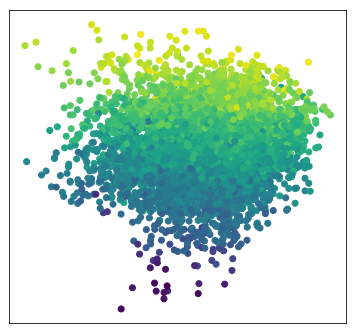}} &
				{\includegraphics[width = 0.7in]{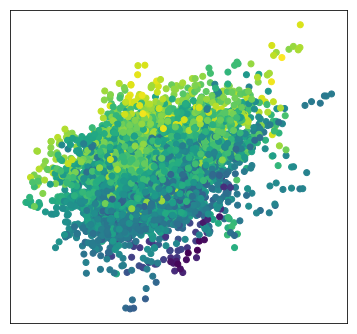}}
			\end{tabular}
			\caption{\label{subfig:vizdoom_latent_corr}}
		\end{subfigure}
		\begin{subfigure}[h]{5cm}
			\centering
			{\includegraphics[width = 1.3in]{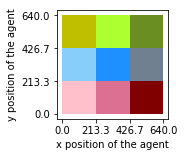}}
			\caption{\label{subfig:vizdoom_latent_walk}}
		\end{subfigure}
		\caption{\ref{subfig:vizdoom_latent_corr} We color the two most correlated latent dimensions learned in the Vizdoom Environment by ground truth values - position and velocity using the checkerboard pattern in \ref{subfig:vizdoom_latent_walk} in the first row, and the ball's velocities in the x and y directions in the second and third rows. Note that the \ac{SLDS} model with $\beta=1$ is unable to learn much about the agent's velocity despite the additional regularization via parameter sharing.}
		\label{fig:vizdoom_latents}
	\end{figure}

\begin{figure}[h!]
		\centering
		\begin{tabular}{c}
			
			\midrule
			\multicolumn{1}{c}{\textbf{DVBF-SLDS}} \\
			\midrule
			{\includegraphics[width = 4in]{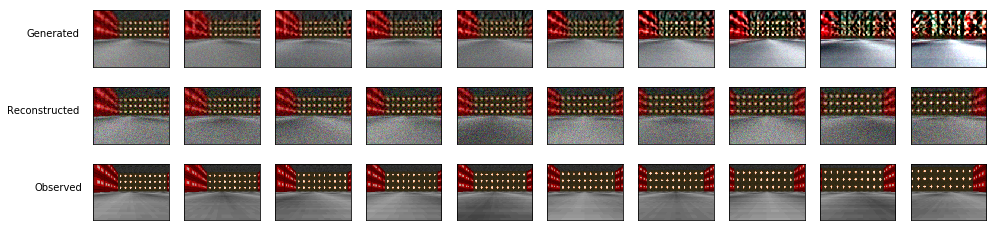}}\\
			\midrule
			\multicolumn{1}{c}{\textbf{DVBF-SLDS-$\beta$}} \\
			\midrule
			{\includegraphics[width = 4in]{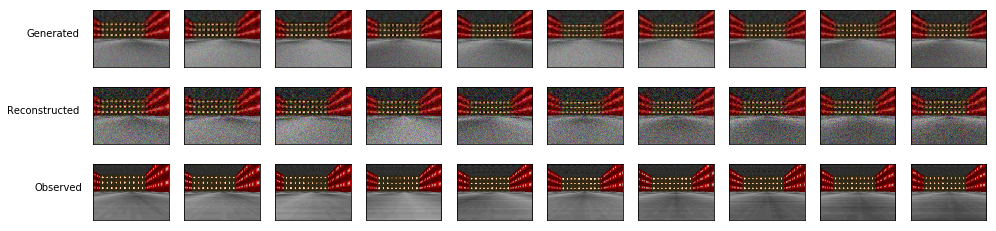}}\\
			\midrule
			\multicolumn{1}{c}{\textbf{DVBF-Non-Shared-$\beta$}} \\
			\midrule
			{\includegraphics[width = 4in]{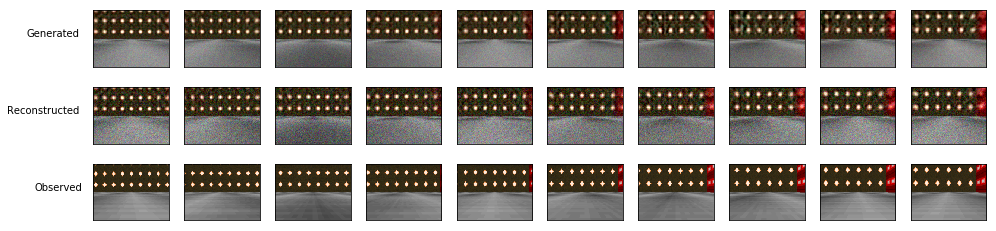}}\\
			\bottomrule
			
		\end{tabular}
		\caption{We show the generated, reconstructed and original trajectories obtained in the Vizdoom Environment for the various models described in table \ref{tab:vizdoom_models} When $\beta$ equals $1$, the model generates comparatively bad image sequences. Note that we can reasonably attribute the poor quality of generated images to an inadequate generative model instead of a bad decoder since the reconstructed images for the same model are sufficiently good.}
		\label{fig:vizdoom_results}
	\end{figure}

	As indicated, a marked difference from the previous experiment is that the images processed here are very high-dimensional ($96 \times 128 \times 3$) compared to the latents ( with dimensions $128$). Consequently, a high $\beta$ value becomes particularly significant for learning transition dynamics in this setting as evident from Table \ref{tab:vizdoom_metrics} and Figures \ref{fig:vizdoom_latents} and \ref{fig:vizdoom_results}, \textbf{even} when (a) parameter-sharing and (b) structured transitions like \ac{SLDS} are used. As before, with an appropriately high beta, the model works well even when (a) and (b) are absent.
	
	\begin{table}[h!]
		\centering
		\begin{center}
			\begin{tabular}{p{3.1cm} p{0.9cm} p{0.7cm} p{1.2cm} p{1.6cm} p{1.7cm} p{0.7cm} p{0.7cm} p{0.7cm}}
				\toprule
				\bf Model & \bf $\mathcal{L}_{ELBO}$ & \bf KL & \bf NLL & \multicolumn{2}{c}{\bf Correlation} & \multicolumn{3}{c}{\bf MSE (For Steps)} \\ 
				& & & & \bf Pos. (x,y) & \bf Vel. (vx,vy)  & \bf 1 & \bf 5  & \bf 10\\ 
				\midrule
				DVBF-SLDS & $\mathbf{1.04e5}$ & ${302.0}$ & $\mathbf{-1.04e5}$ & $0.85$, ${0.88}$ & ${0.10}$, ${0.10}$ & $\mathbf{319.2}$ & ${694.6}$ & $1294$\\
				DVBF-SLDS-$\beta$ & ${9.87e4}$ & $\mathbf{3.385}$ & ${-9.69e4}$ & ${0.93}$, $0.93$ & $\mathbf{0.75}$, $\mathbf{0.90}$ & ${356.2}$ & $\mathbf{429.2}$ & $611.7$\\
				DVBF-Non-Shared-$\beta$ & ${9.73e4}$ & $3.985$ & ${-9.87e4}$ & $\mathbf{0.98}$, $\mathbf{0.97}$ & $0.65$, $0.61$ & ${374.2}$ & $454.5$ & $\mathbf{591.6}$\\
				\bottomrule
			\end{tabular}
		\end{center}
		\caption{\label{tab:vizdoom_metrics} The  different metrics evaluated on   models from \ref{tab:vizdoom_models} for the Vizdoom Environment.}
	\end{table}

	\clearpage
	\pagebreak
	
	\bibliography{bib}

\begin{thebibliography}{10}\itemsep=-1pt

\bibitem{bayer2014learning}
J.~Bayer and C.~Osendorfer.
\newblock Learning stochastic recurrent networks.
\newblock {\em arXiv preprint arXiv:1411.7610}, 2014.

\bibitem{becker2018switching}
P.~Becker-Ehmck, J.~Peters, and P.~Van Der~Smagt.
\newblock Switching linear dynamics for variational {B}ayes filtering.
\newblock In K.~Chaudhuri and R.~Salakhutdinov, editors, {\em Proceedings of
  the 36th International Conference on Machine Learning}, volume~97 of {\em
  Proceedings of Machine Learning Research}, pages 553--562, Long Beach,
  California, USA, 09--15 Jun 2019. PMLR.

\bibitem{chung2015recurrent}
J.~Chung, K.~Kastner, L.~Dinh, K.~Goel, A.~C. Courville, and Y.~Bengio.
\newblock A recurrent latent variable model for sequential data.
\newblock In {\em Advances in neural information processing systems}, pages
  2980--2988, 2015.

\bibitem{NehaDas:Thesis:2019}
N.~Das.
\newblock {Learning State-Space Models of Camera-Based Robots for Intrinsically
  Motivated Control}.
\newblock Master's thesis, Technical University of Munich, Germany, 2019.

\bibitem{deisenroth2011pilco}
M.~Deisenroth and C.~E. Rasmussen.
\newblock Pilco: A model-based and data-efficient approach to policy search.
\newblock In {\em Proceedings of the 28th International Conference on machine
  learning (ICML-11)}, pages 465--472, 2011.

\bibitem{denton2018stochastic}
E.~Denton and R.~Fergus.
\newblock Stochastic video generation with a learned prior.
\newblock {\em arXiv preprint arXiv:1802.07687}, 2018.

\bibitem{fraccaro2017disentangled}
M.~Fraccaro, S.~Kamronn, U.~Paquet, and O.~Winther.
\newblock A disentangled recognition and nonlinear dynamics model for
  unsupervised learning.
\newblock In {\em Advances in Neural Information Processing Systems}, pages
  3601--3610, 2017.

\bibitem{higgins2017beta}
I.~Higgins, L.~Matthey, A.~Pal, C.~Burgess, X.~Glorot, M.~Botvinick,
  S.~Mohamed, and A.~Lerchner.
\newblock beta-vae: Learning basic visual concepts with a constrained
  variational framework.
\newblock In {\em International Conference on Learning Representations}, 2017.

\bibitem{jayaraman2018time}
D.~Jayaraman, F.~Ebert, A.~A. Efros, and S.~Levine.
\newblock Time-agnostic prediction: Predicting predictable video frames.
\newblock {\em arXiv preprint arXiv:1808.07784}, 2018.

\bibitem{karl2016deep}
M.~Karl, M.~Soelch, J.~Bayer, and P.~van~der Smagt.
\newblock Deep variational bayes filters: Unsupervised learning of state space
  models from raw data.
\newblock In {\em Proceedings of the International Conference on Learning
  Representations (ICLR)}, 2017.

\bibitem{karl2017unsupervised}
M.~Karl, M.~Soelch, P.~Becker-Ehmck, D.~Benbouzid, P.~van~der Smagt, and
  J.~Bayer.
\newblock Unsupervised real-time control through variational empowerment.
\newblock {\em arXiv preprint arXiv:1710.05101}, 2017.

\bibitem{klyubin2005empowerment}
A.~S. Klyubin, D.~Polani, and C.~L. Nehaniv.
\newblock Empowerment: A universal agent-centric measure of control.
\newblock In {\em 2005 IEEE Congress on Evolutionary Computation}, volume~1,
  pages 128--135. IEEE, 2005.

\bibitem{2015deepkalmanfilters}
R.~G. {Krishnan}, U.~{Shalit}, and D.~{Sontag}.
\newblock {Deep Kalman Filters}.
\newblock {\em arXiv e-prints}, page arXiv:1511.05121, Nov 2015.

\bibitem{lee2018stochastic}
A.~X. Lee, R.~Zhang, F.~Ebert, P.~Abbeel, C.~Finn, and S.~Levine.
\newblock Stochastic adversarial video prediction.
\newblock {\em arXiv preprint arXiv:1804.01523}, 2018.

\bibitem{2015arXiv150908731M}
S.~{Mohamed} and D.~{Jimenez Rezende}.
\newblock {Variational Information Maximisation for Intrinsically Motivated
  Reinforcement Learning}.
\newblock {\em arXiv e-prints}, page arXiv:1509.08731, 2015.

\bibitem{murphy2012machine}
K.~P. Murphy.
\newblock {\em Machine learning: a probabilistic perspective}.
\newblock MIT press, 2012.

\bibitem{rezende2018taming}
D.~J. Rezende and F.~Viola.
\newblock Taming vaes.
\newblock {\em arXiv preprint arXiv:1810.00597}, 2018.

\bibitem{turner2010statistical}
R.~E. Turner.
\newblock {\em Statistical models for natural sounds}.
\newblock PhD thesis, UCL (University College London), 2010.

\bibitem{watter2015embed}
M.~Watter, J.~Springenberg, J.~Boedecker, and M.~Riedmiller.
\newblock Embed to control: A locally linear latent dynamics model for control
  from raw images.
\newblock In {\em Advances in neural information processing systems}, pages
  2746--2754, 2015.

\bibitem{xie2017neural}
Z.~Xie.
\newblock Neural text generation: A practical guide.
\newblock {\em arXiv preprint arXiv:1711.09534}, 2017.

\end{thebibliography}
	\bibliographystyle{bib}
	
	\clearpage
	\appendix
	
	\section{Acronyms}
	\begin{acronym}
		\acro{RL}{Reinforcement Learning}
		\acro{CNN}{Convolutional Neural Network}
		\acro{VAE}{Variational Autoencoder}
		\acro{LSTM}{Long Short-Term Memory}
		\acro{MLP}{Multi-Layered Perceptron}
		\acro{DVBF}{Deep Variational Bayes Filter}
		\acro{SLDS}{Switching Linear Dynamics System}
		\acro{SGD}{Stochastic Gradient Descent}
		\acro{VI}{Variational Inference}
		\acro{SSM}{State Space Model}
		\acro{MI}{Mutual Information}
		\acro{GECO}{Generalized ELBO with Constrained Optimization}
		\acro{KL}{Kullback Leibler}
		\acro{NLL}{Negative Log Likelihood}
	\end{acronym}

	\section{Pendulum Dynamics}\label{app:pend_dynamics}
	The dynamics in the pendulum experiment are generated using the following equation:
	
	\begin{equation*}
	ml^2\ddot{\psi}(t) = -\mu \dot{\psi}(t) + mgl \sin(\psi(t)) + u(t)
	\end{equation*}
	
	where $m$ denotes the pendulum mass, $l$, the string length and $g$ denotes the acceleration due to gravity. $u(t)$  is the torque control at time $t$ and is chosen freely and determines the value of the pendulum angle $\psi(t)$ along with the constants  $m, l, g$ and $\mu$, which are fixed at $m=l=1$, $g=9.8$, $\mu=0.5$.
	
	\section{Generalized ELBO Constrained Optimization} \label{subsec:GECO}
	
	We additionally complement our work with that of Rezende et al.'s \ac{GECO} loss \cite{rezende2018taming}, that instead of setting $\beta$ as a hyperparameter and arbitrarily deciding the value of the constraint $\epsilon$ on KL Divergence, reformulates the loss as a Lagrangian that constrains the reconstruction error with a hyper-parameter $\kappa$ instead - a more interpretable value:
	
	\begin{equation}
	\lambda \cdot \underset{\rho(x) q(z|x)}{\mathbb{E}}\big[ \mathcal{C}(x,g(z))\big] + \underset{\rho(x)}{\mathbb{E}} \big[KL(q(z|x)||p(z))\big]
	\end{equation}
	
	where $\mathcal{C}(x,g(z)$ is the reconstruction error with an inequality constraint such that:    
	\begin{equation}
	\mathcal{C}(x,g(z) = ||(x-g(z))||^2 - \kappa^2 \geq 0
	\end{equation}
	
	Note that in addition to the generative and inference parameters, also optimize for the Lagrange multiplier $\lambda$. Consequently, we not only ease up the process of hyper-parameter tuning since $\kappa$ is easier to set due its direct correlation with the desired reconstruction quality, but also we now no longer need to anneal the training process. 
	
	\subsection{Bouncing Ball in a Box} \label{sec:dvbf_bb}
	
	This Dataset again comprises of sequences of image and control vector pairs sampled from the Bouncing Ball Environment - a simulation of a spherical mass rolling on a plane surface surrounded by walls. We use pybox2d for simulating the environment dynamics for this experiment. The control vector (chosen freely at each time for data generation) is made up of forces in x and y directions. This implies that the environment dynamics are non-Markovian and non-Linear with respect to the position and velocity of the agent - quantites that can fully specify the system state at any given point.
	
	\begin{table}[h!]
		\centering
		\begin{center}
			\begin{tabular}{l p{3cm} p{3cm} p{0.5cm} p{2.5cm} l}
				\toprule
				\bf Model & \bf Transition Network & \bf Shared Prior and Posterior parameters & \bf $\beta$ & \bf Annealing Temperature & \bf $\kappa$\\ 
				\midrule
				DVBF-SLDS & \ac{SLDS} & mean $\mu_{trans}$ & $1.0$ & $1.0$ & N.A\\
				DVBF-Non-Shared & \ac{MLP} & None & $1.0$ & $1.0$ & N.A\\
				DVBF-Non-Shared-$\beta$ & \ac{MLP} & None & $16.0$ & $20e3$ & N.A\\
				DVBF-GECO & \ac{MLP} & None & N.A & N.A & $3.0$\\
				\bottomrule
			\end{tabular}
		\end{center}
		\caption{\label{tab:bb_models} Model Description for the Bouncing Ball Environment. The first two models optimize the $\mathcal{L}_{ELBO}$ value for sequential data - equation \ref{eq:seq_ELBO}. DVBF-Non-Shared-$\beta$ maximizes a $\beta$ dependent lower bound to $\mathcal{L}_{ELBO}$ described in equation \ref{eq:LELBO_beta}. Finally, DVBF-GECO optimizes a sequential version of Generalized ELBO with Constraints \cite{rezende2018taming}. More details about this loss is provided in Section \ref{sec:Method}. }
	\end{table}
	
	\begin{table}[h!]
		\centering
		\begin{center}
			\begin{tabular}{p{2.5cm} l p{1cm} p{1.5cm} p{1cm} p{1cm} p{1cm} p{1cm} p{1cm} p{1cm}}
				\toprule
				\bf Model & \bf $\mathcal{L}_{ELBO}$ & \bf KL & \bf NLL & \multicolumn{2}{c}{\bf Correlation} & \multicolumn{3}{c}{\bf MSE (For Steps)} &  \bf Iter.\\ 
				& & & & \bf Postion x,y & \bf Velocity vx,vy  & \bf 1 & \bf 5  & \bf 10 & (in 1000s.)\\ 
				\midrule
				DVBF-SLDS & $6748$ & $\mathbf{5.406}$ & $-6764$ & $0.92$ ${0.97}$ & ${0.55}$ $\mathbf{0.64}$ & ${12.52}$ & $\mathbf{13.56}$ & $\mathbf{15.11}$ & 120\\
				DVBF-Non-Shared & $\mathbf{7069}$ & $10.97$ & $\mathbf{-7063}$ & ${0.93}$ $0.84$ & ${0.55}$ $0.40$ & $\mathbf{11.29}$ & $14.03$ & $16.42$ & 550\\
				DVBF-Non-Shared-$\beta$ & ${6710}$ & $1.78$ & ${-6714}$ & $\mathbf{0.99}$ $\mathbf{0.99}$ & $\mathbf{0.67}$ $0.55$ & ${12.72}$ & $14.06$ & $15.74$ & 120\\
				\bottomrule
			\end{tabular}
		\end{center}
		
		\begin{center}
			\begin{tabular}{p{2.5cm} p{1cm} p{1cm} p{1.5cm} p{1cm} p{1cm} p{1cm} p{1cm} p{1cm} p{1cm}}
				\toprule
				\bf Model & \bf Constr. & \bf KL & \bf $\mathcal{C}(x, g(z))$ & \multicolumn{2}{c}{\bf Correlation} & \multicolumn{3}{c}{\bf MSE (For Steps)} &  \bf Iter.\\ 
				& $\mathcal{L}_{ELBO}$ & & $\kappa = 3.0$ & \bf Position x,y & \bf Velocity vx,vy  & \bf 1 & \bf 5  & \bf 10 & (in 1000s.)\\ 
				\midrule
				DVBF-GECO & $-0.56$ & ${0.712}$ & $-0.023$ & $0.78$ $0.88$ & $0.49$ $0.60$ & ${12.71}$ & ${14.80}$ & ${16.22}$ & 120\\
				\bottomrule
			\end{tabular}
		\end{center}
		\caption{\label{tab:bb_metrics} We compare here, several different metrics for models described in \ref{tab:bb_models} for the Bouncing Ball Environment}
	\end{table}
	
\begin{figure}[h!]
		\centering
		\begin{subfigure}[h]{10cm}
			\begin{tabular}{c p{1.3cm} p{1.3cm} p{1.3cm} p{1.3cm}}
				\toprule
				&\bf DVBF-SLDS & \bf DVBF-Non-Shared & \bf DVBF-Non-Shared-$\beta$ & \bf DVBF-GECO\\ 
				\midrule
				\raisebox{-.5\normalbaselineskip}[0pt][0pt]{\rotatebox[origin=c]{90}{\hspace{2cm}\textbf{Latent Walk}}} &
				{\includegraphics[width = 0.7in]{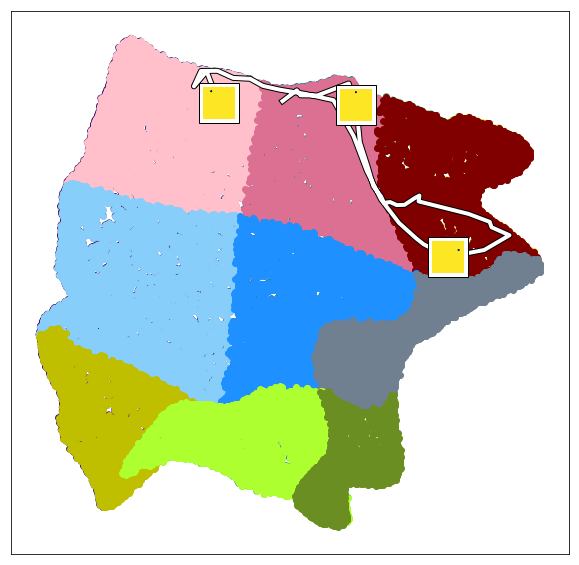}} &
				{\includegraphics[width = 0.7in]{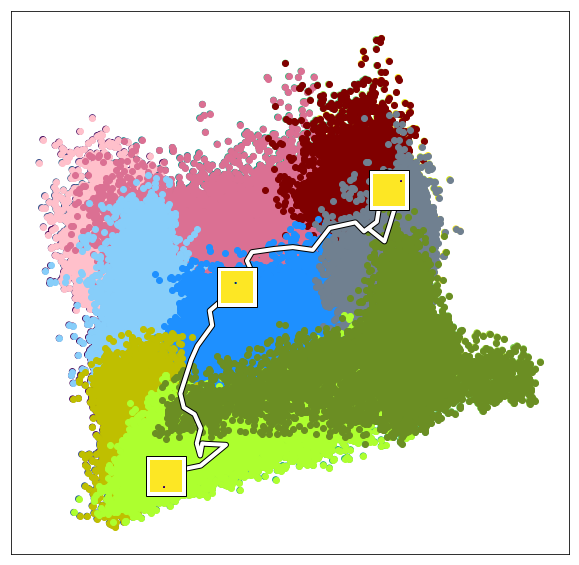}} &
				{\includegraphics[width = 0.7in]{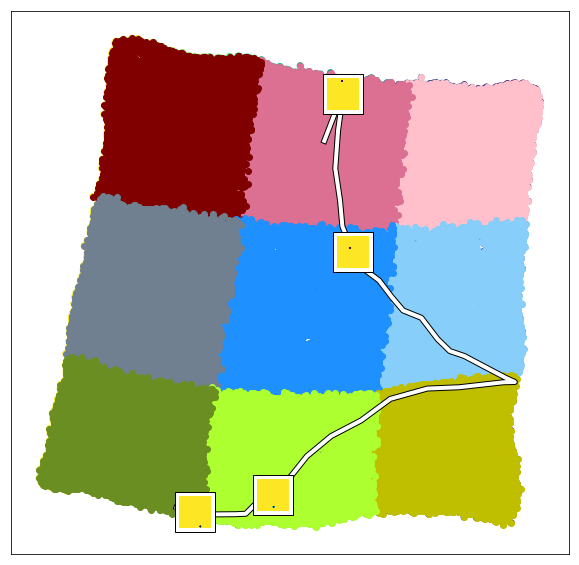}} &
				{\includegraphics[width = 0.7in]{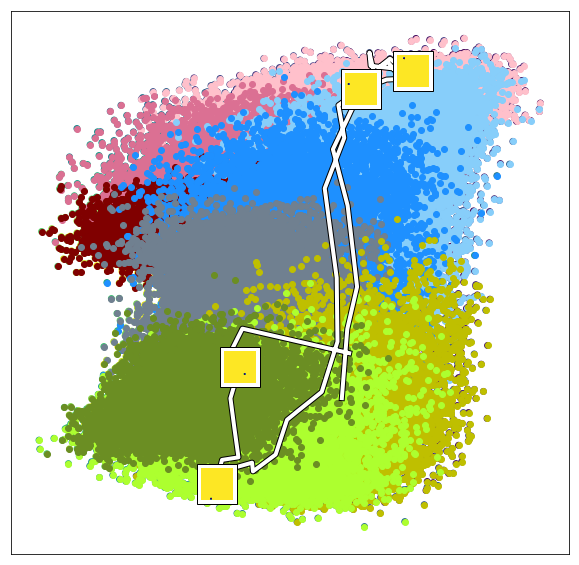}}\\
				\begin{turn}{90}
					\bf Velocity(x)
				\end{turn} &
				{\includegraphics[width = 0.7in]{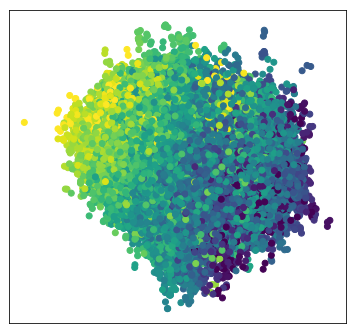}} &
				{\includegraphics[width = 0.7in]{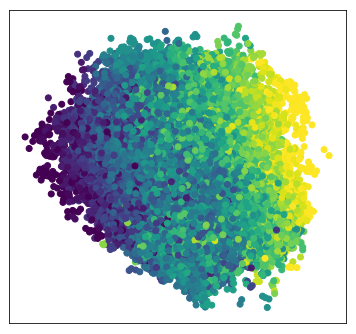}} &
				{\includegraphics[width = 0.7in]{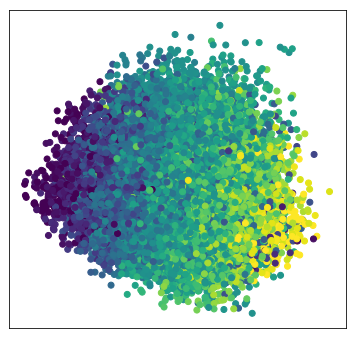}} &
				{\includegraphics[width = 0.7in]{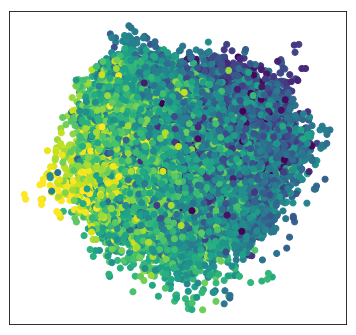}}\\
				\begin{turn}{90}
					\bf Velocity(y)
				\end{turn} &
				{\includegraphics[width = 0.7in]{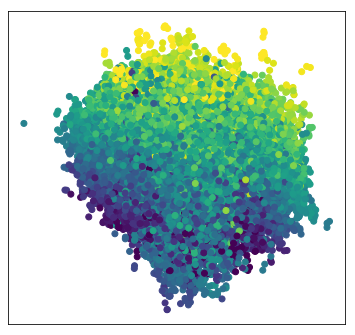}} &
				{\includegraphics[width = 0.7in]{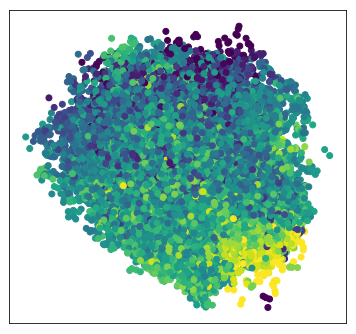}} &
				{\includegraphics[width = 0.7in]{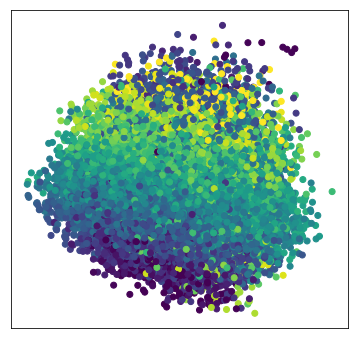}} &
				{\includegraphics[width = 0.7in]{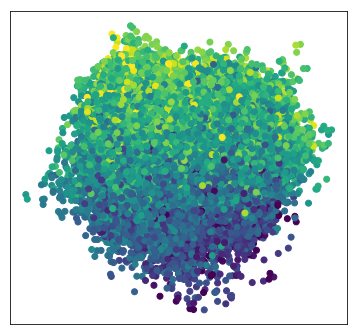}}
			\end{tabular}
			\caption{\label{subfig:bb_latent_corr}}
		\end{subfigure} 
		\begin{subfigure}[h]{3cm}
			\centering
			{\includegraphics[width = 1.3in]{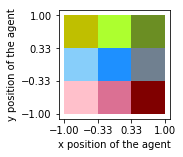}}
			\caption{\label{subfig:bb_latent_walk}}
		\end{subfigure}
		\caption{\ref{subfig:bb_latent_corr} We color the two most correlated latent dimensions learned in the Bouncing Ball Environment by ground truth values - position and velocity using the checkerboard pattern in \ref{subfig:bb_latent_walk} in the first row, and the ball's velocities in the x and y directions in the second and third rows.}
		\label{fig:bb_latents}
	\end{figure}

	We evaluate four different models (Table \ref{tab:bb_models}) for this environment with DVBF-SLDS as our baseline, and once more show via several evaluation metrics (Table \ref{tab:bb_metrics}) and plots correlating latents to the ground truth values of the system state (Figure \ref{fig:bb_latents}) that decomposing the approximate posterior provides the key incentive in learning a adequately good generative model. Note however, that converging to this model (DVBF-Non-Shared) without parameter sharing and structured transition networks like switching linear dynamics. A considerable reduction in convergence time was observed when an appropriate value of $\beta$ is used to the train the model in conjunction with an annealing scheme. Similar results are obtained when we use the \ac{GECO} model (See section \ref{subsec:GECO}) (see Table \ref{tab:bb_metrics}), with slightly less effort expended in hyper-parameter tuning.

\begin{figure}[h!]
		\centering
		\begin{tabular}{c}
			
			\midrule
			\multicolumn{1}{c}{\textbf{DVBF-SLDS}} \\
			\midrule
			{\includegraphics[width = 4in]{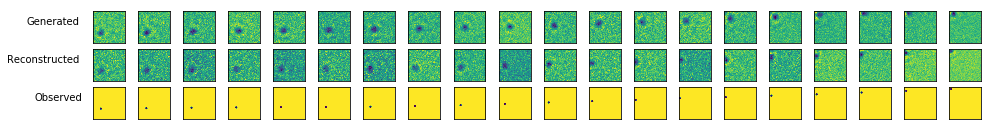}}\\
			\midrule
			\multicolumn{1}{c}{\textbf{DVBF-Non-Shared}} \\
			\midrule
			{\includegraphics[width = 4in]{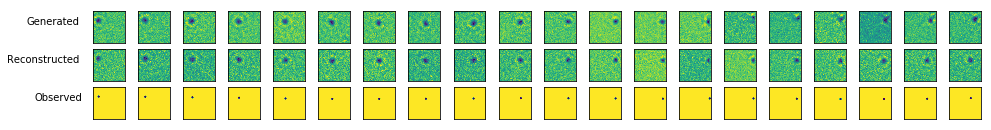}}\\
			\midrule
			\multicolumn{1}{c}{\textbf{DVBF-Non-Shared-$\beta$}} \\
			\midrule
			{\includegraphics[width = 4in]{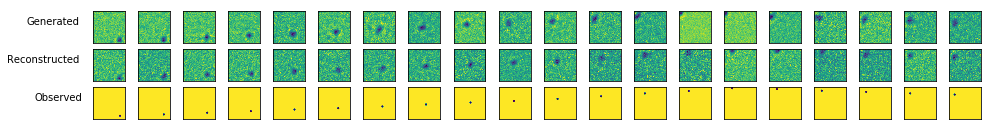}}\\
			\midrule
			\multicolumn{1}{c}{\textbf{DVBF-GECO}} \\
			\midrule
			{\includegraphics[width = 4in]{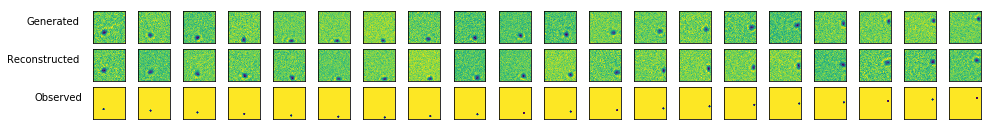}}\\
			\bottomrule
		\end{tabular}
		\caption{We show the generated, reconstructed and original trajectories obtained in the Bouncing Ball Environment for the various models described in table \ref{tab:bb_models}}
		\label{fig:bb_results}
	\end{figure}

	Finally we show the original, reconstructed and generated samples for the models specified in Table \ref{tab:bb_models} in Figure \ref{fig:bb_results}

	\clearpage
	
	\section{Empowerment}
	
	In this section we verify the reliability of the model learnt from high dimensional observations using a high $\beta$ value for an intrinsically motivated control task. The intrinsic motivation measure used here is called Empowerment \cite{klyubin2005empowerment} and is defined as the channel capacity between the current action and the resultant state of the environment:
	
	\begin{equation}
	\mathcal{E}(z) = \max_{\omega(u|z)} I(z'; u|z)
	\label{eq:empowerment}
	\end{equation}
	
	where $\omega$ denotes the set of probability densities with which an action $u$ is picked. To obtain empowerment we pick the probability density that maximizes the mutual information $I(z'; u|z)$, where 
	
	\begin{equation}
	\begin{split}
	I(z'; u|z) &= \int p(z', u|z) \log \frac{ p(z', u|z)}{p(z'|z)\omega(u|z)} du dz'
	\end{split}
	\end{equation}
	
	More intuitively, empowerment is indicative of the number of future possibilities that are available at a particular state $z_t$. 
	The motivation for maximizing this quantity is simply to keep moving to a state from where maximum possible options are accessible, i.e a state where the agent is maximally prepared. For a more in-depth discussion of the philosophy behind empowerment and empowerment-based control, look at \cite{klyubin2005empowerment, karl2017unsupervised}
	
	Empowerment is rather difficult to estimate since the mutual information term in equation \ref{eq:empowerment} is intractable for continuous state or action spaces. A way to get around this predicament is to approximate the mutual information term with the Barber-Agakov Bound, and thus obtaining the following lower-bound for empowerment:
	
	\begin{equation}
	\hat{\mathcal{E}}(z) = \max_{\omega(u|z)} \hat{I}(z'; u|z) = \max_{\omega(u|z)} \underset{z' \sim p(z'|u,z)}{\underset{u \sim \omega(u|z)}{\mathbb{E}}} \big[\log{q(u|z,z')} - \log{\omega(u|z)}\big] \leq {\mathcal{E}}(z)
	\label{eq:lower_bound_empowerment}
	\end{equation}
	
	Mohamed et al. \cite{2015arXiv150908731M} use a model free approach for optimizing this bound. However, since lacking a differentiable model of dynamics, this basically boils down to this essentially boils down to maximizing the entropy of the source distribution $H(\omega) = \underset{u \sim \omega(u|z)}{\mathbb{E}}\big[ - \log{\omega(u|z)}\big]$ in an unconstrained manner, they needed additional measures in place to ensure that doing this does not degenerate to a diverging process. 
	
	\cite{karl2017unsupervised}, on the other hand use \ac{DVBF} to model the dynamics, thus providing a link to connect the flow of gradients between $\omega$ and $q$ in a stable manner. 
	
	We show results from using $\beta$-DVBF instead of \ac{DVBF} for high-dimensional images obtained from the Vizdoom environment with all else being similar to the experiments performed in \cite{karl2017unsupervised}.
	
\begin{figure}
		\centering
		\begin{tabular}{c}
			\toprule
			\bf Empowerment Map\\ 
			\midrule
			{\includegraphics[width = 2in]{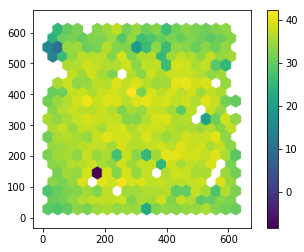}}
		\end{tabular}
		\caption{Vizdoom: This image shows the empowerment values learned corresponding to agent's position in the bounded box. Note that the empowerment values are lower near the edges and corners of the box.}
		\label{fig:vizdoom_emp_map}
	\end{figure}

	As expected and shown in Figure \ref{fig:vizdoom_emp_map}, the empowerment maximizing policy that we learn here moves the agent away from the walls and corners. The fact that we are able to learn such a policy successfully further testifies to the accuracy of the transition model learned from the observation trajectory.
	
\begin{figure}
		\centering
		\begin{subfigure}[h]{15cm}
			\centering
			\begin{tabular}{cc}
				\includegraphics[width = 2.5in]{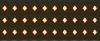}&
				\includegraphics[width = 2.5in]{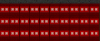}
			\end{tabular}
			\caption{\label{subfig:vizdoom_walls}}
		\end{subfigure}\\
		\begin{subfigure}[h]{15cm}
			\includegraphics[width = 6in]{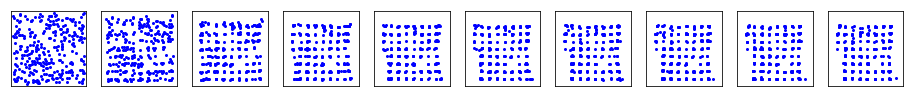}
			\caption{\label{subfig:vizdoom_rollout}}
		\end{subfigure}
		\caption{Vizdoom: In the top row, Figure \ref{subfig:vizdoom_walls} shows the two distinct patterns that provide texture to the walls of the bounding box. Figure \ref{subfig:vizdoom_rollout} shows the rolled out trajectory for 300 vizdoom samples over 80 steps according to an empowerment maximizing policy for 7-step open-loop empowerment. Note that on following this policy, the sampled agents roughly converge to a grid-like pattern reminiscent of the pattern on the walls in \ref{subfig:vizdoom_walls}.}
		\label{fig:vizdoom_rollout}
	\end{figure}

	Figure \ref{fig:vizdoom_rollout} shows the rollout of an empowerment maximizing policy for a 300 samples over 80 steps. Note especially that (a) the empowerment maximizing policy tends to move the agent away from walls and corners, and (b) the samples are eventually arranged in a grid like fashion instead of being uniformly or normally distributed in the central area. We speculate that this might be due to the empowerment maximizing policy driving the agent to \emph{interesting} states, positions that have the potential of informing the agent the most when it takes an action through a greater change in its perceived input. In this case, we believe the agent aligns itself accordingly with the pattern it sees on the walls (See Figure \ref{fig:vizdoom_data}). 
	
	\clearpage
	
	\section{Additional Results for different $\beta$ Configurations} \label{app:beta}
	
	\subsection{Pendulum}
	
	\begin{table}[h]
		\centering
		\begin{center}
			\begin{tabular}{p{1cm} l l l p{1cm} p{1.5cm} p{1cm} p{1cm} p{1cm} p{1cm}}
				\toprule
				\bf $\beta$-value & \bf $\mathcal{L}_{ELBO}$ & \bf KL & \bf NLL & \multicolumn{2}{c}{\bf Correlation} & \multicolumn{3}{c}{\bf MSE (For Steps)} & \bf Temp.\\ 
				& & & & \bf Angle & \bf Velocity  & \bf 1 & \bf 5  & \bf 10 & \\ 
				\midrule
				1.0 & ${746.4}$ & $1.828$ & $-748.2$ & $\mathbf{0.74}$ & $0.73$ & $0.2454$ & $0.4667$ & $0.6414$ & 1 \\
				4.0 & $\mathbf{747.8}$ & ${0.844}$ & $\mathbf{-748.6}$ & $\mathbf{0.74}$ & $\mathbf{0.95}$ & $\mathbf{0.2451}$ & $\mathbf{0.4616}$ & $\mathbf{0.6246}$  & 5e3\\
				8.0 & ${746.1}$ & $\mathbf{0.774}$ & $-746.9$ & ${0.71}$ & $0.88$ & ${0.2460}$ & $0.4662$ & $0.6313$ & 5e3 \\
				\bottomrule
			\end{tabular}
		\end{center}
		\caption{\label{tab:app_metrics} Metrics corresponding to different values of $\beta$ and Annealing Temperature when a \ac{DVBF} model that uses \ac{MLP} transitions without parameter sharing but with decomposed posterior is trained for 60K iterations for the \textbf{Pendulum Environment}. We compare here the evidence lower bound, negative log-likelihood, KL divergence between the prior and posterior, correlation of latents with ground truth quantities and Mean Squared Error for next 1, 5 and 10 predictive steps.}
	\end{table}	
	\subsection{Bouncing Ball}
	
	\begin{table}[h]
		\centering
		\begin{center}
			\begin{tabular}{p{2.5cm} l p{1cm} p{1.5cm} p{1cm} p{1cm} p{1cm} p{1cm} p{1cm} p{1cm}}
				\toprule
				\bf $\beta$-value & \bf $\mathcal{L}_{ELBO}$ & \bf KL & \bf NLL & \multicolumn{2}{c}{\bf Correlation} & \multicolumn{3}{c}{\bf MSE (For Steps)} & \bf Temp.\\ 
				& & & & \bf Postion x,y & \bf Velocity vx,vy  & \bf 1 & \bf 5  & \bf 10 &\\ 
				\midrule
				1.0 & $\mathbf{7042}$ & $21.66$ & $\mathbf{-7063}$ & $0.79$ $0.82$ & $0.22$ $0.28$ & $\mathbf{11.29}$ & ${16.12}$ & $18.64$ & 1.0 \\
				
				4.0& ${6932}$ & $9.23$ & ${-6938}$ & $0.80$ $0.89$ & $0.37$ $0.32$ & ${11.81}$ & $17.30$ & $16.86$ & 20e3 \\
				
				8.0& ${6895}$ & $4.31$ & ${-6897}$ & $0.87$ $0.79$ & $0.64$ $0.53$ & ${11.99}$ & $14.41$ & $16.64$ & 20e3 \\
				16.0& ${6710}$ & $\mathbf{1.78}$ & ${-6714}$ & $\mathbf{0.99}$ $\mathbf{0.99}$ & $\mathbf{0.67}$ $\mathbf{0.55}$ & ${12.72}$ & $\mathbf{14.06}$ & $\mathbf{15.74}$ & 20e3 \\
				\bottomrule
			\end{tabular}
		\end{center}
		\caption{\label{tab:app_bb_metrics} Metrics corresponding to different values of $\beta$ and Annealing Temperature when a \ac{DVBF} model that uses \ac{MLP} transitions without parameter sharing but with decomposed posterior is trained for 120K iterations for the \textbf{Bouncing Ball Environment}. We compare here the evidence lower bound, negative log-likelihood, KL divergence between the prior and posterior, correlation of latents with ground truth quantities and Mean Squared Error for next 1, 5 and 10 predictive steps.}
	\end{table}

	\section{Architectural details regarding the Experiments}
	
	\subsection{Pendulum}
	
	The pendulum dynamics were simulated by numerically integrating equation:
	
	\begin{equation*}
	ml^2\ddot{\psi}(t) = -\mu \dot{\psi}(t) + mgl \sin(\psi(t)) + u(t)
	\end{equation*}
	
	to obtain angles and angular velocities. The angles at each time step were plotted on a 16x16 plot using a Gaussian distribution over the weight's position.
	
	\begin{itemize}
		\item \textbf{Observation Size ($n_o$): }$16\times16$
		\item \textbf{Number of Actions ($n_u$): }$1$
		\item \textbf{Number of Latent Dimensions ($n_z$): }$64$
		\item \textbf{Inverse Measurement Model: } A 4-layer convolution network with 4, 8, 16 and 64 filters with size $3\times3$ + Dense layer (256 units, ReLu activation) + (Dense Layer ($n_z$ units), Dense Layer ($n_z$ units))
		\item \textbf{Decoder: } Dense layer (256 units, ReLu activation) + Dense Layer (64 units) + A 4-layer transposed convolution network with 16, 8, 4 and 1 filters with size $3\times3$ + (Dense Layer($n_o$ units), 1 Variable)
		\item \textbf{Prior and Posterior Transition Models: }  Dense layer (256 units, sigmoid activation)
	\end{itemize}
	
	\subsection{Bouncing Ball}
	
	The dynamics of this environment were simulated using a PyBox2D engine and rendered using OpenCV.
	
	\begin{itemize}
		\item \textbf{Observation Size ($n_o$): }$64\times64$
		\item \textbf{Number of Actions ($n_u$): }$2$
		\item \textbf{Number of Latent Dimensions ($n_z$): }$64$
		\item \textbf{Inverse Measurement Model: } A 6-layer convolution network with 4, 8, 16, 32, 64 and 128 filters with size $3\times3$ + Dense layer (256 units, ReLu activation) + (Dense Layer ($n_z$ units), Dense Layer ($n_z$ units))
		\item \textbf{Decoder: } Dense layer (256 units, ReLu activation) + Dense Layer (128 units) + A 6-layer transposed convolution network with 64, 32, 16, 8, 4 and 1 filters with size $3\times3$ + (Dense Layer ($n_o$ units), 1 Variable)
		\item \textbf{Prior and Posterior Transition Models: }  Dense layer (256 units, sigmoid activation)
	\end{itemize}
	
	\subsection{Vizdoom}
	
	The dynamics of this environment were simulated using a PyBox2D engine and rendered using the Vizdoom renderer.
	
	\begin{itemize}
		\item \textbf{Observation Size ($n_o$): }$96\times128\times3$
		\item \textbf{Number of Actions ($n_u$): }$2$
		\item \textbf{Number of Latent Dimensions ($n_z$): }$128$
		\item \textbf{Inverse Measurement Model: } A 6-layer convolution network with 8, 16, 32, 64, 128 and 256 filters with size $3\times3$ + $2\times$Dense layer (256 units, ReLu activation) + (Dense Layer ($n_z$ units), Dense Layer ($n_z$ units))
		\item \textbf{Decoder: } Dense layer (256 units, ReLu activation) + $2\times$Dense layer (256 units, ReLu activation) + A 6-layer transposed convolution network with 128, 64, 32, 16, 8 and 3 filters with size $3\times3$ + (Dense Layer ($n_o$ units), 1 Variable)
		\item \textbf{Prior and Posterior Transition Models: }  Dense layer (256 units, sigmoid activation)
	\end{itemize}
	
\end{document}